\documentclass[conference]{IEEEtran}
\IEEEoverridecommandlockouts
\usepackage{cite}
\usepackage{amsmath,amssymb,amsfonts}

\usepackage{algorithmic}
\usepackage{graphicx}
\usepackage{textcomp}
\usepackage{xcolor}
\def\BibTeX{{\rm B\kern-.05em{\sc i\kern-.025em b}\kern-.08em
    T\kern-.1667em\lower.7ex\hbox{E}\kern-.125emX}}
\begin{document}

\title{\textit{INTERACTION}: A Generative XAI Framework for Natural Language Inference Explanations\\

}

\author{\IEEEauthorblockN{Jialin Yu\IEEEauthorrefmark{1},
Alexandra I. Cristea\IEEEauthorrefmark{2}, Anoushka Harit\IEEEauthorrefmark{3} Zhongtian Sun\IEEEauthorrefmark{4}}
\IEEEauthorblockN{Olanrewaju Tahir Aduragba\IEEEauthorrefmark{5},
Lei Shi\IEEEauthorrefmark{6} and
Noura Al Moubayed\IEEEauthorrefmark{7}}
\IEEEauthorblockA{Department of Computer Science,
Durham University\\
Durham, UK\\
Email: \{\IEEEauthorrefmark{1}jialin.yu,
\IEEEauthorrefmark{2}alexandra.i.cristea,
\IEEEauthorrefmark{3}anoushka.harit,
\IEEEauthorrefmark{4}	zhongtian.sun\}@durham.ac.uk \\ \{\IEEEauthorrefmark{5}olanrewaju.m.aduragba, \IEEEauthorrefmark{6}lei.shi, \IEEEauthorrefmark{7}noura.al-moubayed\}@durham.ac.uk}}


\maketitle

\begin{abstract}
\textit{XAI with natural language processing} aims to produce human-readable explanations as evidence for AI decision-making, which addresses explainability and transparency. However, from an HCI perspective, the current approaches only focus on delivering a single explanation, which fails to account for the \textit{diversity of human thoughts and experiences} in language. This paper thus addresses this gap, by proposing a generative XAI framework, \textit{INTERACTION} ({expla\underline{\textbf{I}}n a\underline{\textbf{N}}d predic\underline{\textbf{T}} th\underline{\textbf{E}}n que\underline{\textbf{R}}y with contextu\underline{\textbf{A}}l \underline{\textbf{C}}ondi\underline{\textbf{T}}ional var\underline{\textbf{I}}ational aut\underline{\textbf{O}}-e\underline{\textbf{N}}coder}). Our novel framework presents explanation in two steps: (\textit{step one}) Explanation and Label Prediction; and (\textit{step two}) Diverse Evidence Generation. We conduct intensive experiments with the Transformer architecture on a benchmark dataset, e-SNLI \cite{camburu2018snli}. Our method achieves competitive or better performance against state-of-the-art baseline models on explanation generation (up to 4.7\% gain in BLEU) and prediction (up to 4.4\% gain in accuracy) in \textit{step one}; it can also generate multiple diverse explanations in \textit{step two}.  
\end{abstract}


\section{Introduction}
\label{sec: introduction}


Traditionally, natural language processing (NLP) applications are built based on techniques that are inherently more explainable. Examples of such techniques are often referred to as \textit{`white box'} techniques, including rule-based heuristic systems, decision trees, hidden Markov models, etc. In recent years, due to the advancement of deep learning, a \textit{`black box'} technique, deep neural network, has become the dominant approach \cite{danilevsky2020survey}. With the advancement of deep neural networks, their ubiquitousness comes at the expense of less interpretability. Hence, concerns have been raised on whether deep neural networks can make reasonable judgements \cite{mcallister2017concrete,challen2019artificial}, which further triggers an interest in explainable artificial intelligence (XAI) research \cite{arrieta2020explainable}.

With XAI techniques in NLP applications, researchers first focused on \textit{feature}-based \cite{voskarides2015learning,godin2018explaining}, \textit{model}-based \cite{ribeiro2016should}, and \textit{example}-based \cite{croce2019auditing} explanation techniques. However, even for experts working as data scientists in the industry, interpreting results from these models was found to be hard and bias-prone \cite{kaur2020interpreting}. To reduce human interpretation bias, directly generating natural language explanations seemed a better medium for presentation. Rather than based on carefully designed additional tools, \textit{XAI with natural language} produced human-readable explanations as evidence for AI decision-making \cite{wiegreffe2021teach}.

The current state-of-the-art approaches, such as those in \cite{narang2020wt5, kumar2020nile}, are limited by presenting a single explanation only. However, from an HCI research perspective, it is hard to account for the diversity of human thoughts and experience \cite{aroyo2015truth}. Indeed, natural language allows expressing the same semantic content in various `correct' (i.e., semantically similar) forms, subject to cognitive biases, social expectations, and socio-cultural backgrounds \cite{miller2019explanation}.

This paper addresses this gap by, proposing a generative XAI framework, which presents explanations in two steps: (step one) Explanation and Label Prediction, and (step two) Diverse Evidence Generation. In step one, we offer the most probable explanation and label prediction, similar to other prior work in literature \cite{narang2020wt5, kumar2020nile}. In our original step two, we adopt deep generative models, to generate multiple diverse explanations via posterior analysis in the latent space. We evaluate our method specifically on an natural language inference (NLI) task \cite{bowman2015large}, which determines whether a `hypothesis' is true (entailment), false (contradiction), or undetermined (neutral), given a `premise'. To perform this, an appropriate dataset is needed. Current NLI datasets, however, contain annotation artefacts, which allow the models to make predictions based on spurious correlations \cite{gururangan2018annotation}. To address annotation artefacts in data, Camburu \textit{et al.} \cite{camburu2018snli} suggest that spurious correlations are much harder to be captured with natural language explanations and propose a large-scale benchmark dataset (e-SNLI), which contains NLI data points and their associated explanations. In this paper, we present our studies thus on this dataset, with the Transformer architecture, as further explained in Section \ref{sec: preliminary-experiments} and Section \ref{sec: esnli-main}.

Our main contributions include: \textit{(i)} a novel two-step generative XAI framework, \textbf{INTERACTION}, which presents explanations in two steps: (step one) Explanation and Label Prediction; and (step two) Diverse Evidence Generation; \textit{(ii)} the first study on spurious correlation on the e-SNLI dataset with Transformer architecture; \textit{(iii)} demonstrating the benefits of our framework, \textbf{INTERACTION}, against state-of-the-art baseline models with empirical experiments; and \textit{(iv)} a solid deep generative model baseline for future research in the XAI field.

\section{Related Work}

\subsection{Explainable Artificial Intelligence for Natural Language Processing}

General XAI approaches can be categorised in two main ways: \cite{guidotti2018survey,tjoa2020survey}: 1) Local vs Global, and 2) Self-Explaining vs Post-Hoc. Our work contributes to explainable artificial intelligence (XAI) from two perspectives: \textit{Local} and \textit{Self-Explaining}, as we provide explanations based on fine-granularity individual input, and our explanations are directly interpretable.

In terms of explanation techniques and their applications to NLP there are, in general, five different types \cite{danilevsky2020survey}: 1) feature importance, 2) surrogate model, 3) example-driven, 4) provenance-based, and 5) declarative induction. The first three are more widely adopted and have already been described briefly in section \ref{sec: introduction}. The provenance-based technique refers to visualising some or all of the prediction process, such as in \cite{zhou2018interpretable, amini2019mathqa}. Our work uses the \textit{declarative induction technique}, which tackles the challenging task of providing human-readable representations as part of the results, such as in \cite{camburu2018snli, prollochs2019learning}. Our work further extends \cite{camburu2018snli} with a \textit{probabilistic treatment}. 


\subsection{Supervised Deep Generative Models for Natural Language Processing}

Our work is associated with deep generative models, which is based on Neural Variational Inference (NVI) \cite{kingma2013auto,mnih2014neural,rezende2014stochastic}. NVI is also known as amortised variational inference in the literature and can be considered as an extension of the mean-field variational inference \cite{jordan1999introduction, bishop2006pattern}. The NVI technique uses data-driven neural networks instead of more restrictive statistical inference techniques. NVI allows us to infer unobservable latent random variables that generate the observed data and are thus very efficient for data with hidden structures, such as natural language.

NVI has been successfully applied in various NLP applications including topic modelling \cite{miao2016neural, srivastava2017autoencoding}, machine translation \cite{su2018variational, pagnoni2018conditional}, text classification \cite{miao2016neural}, conversation generation \cite{zhao2017learning,gao2019discrete}, and story generation \cite{fang2021transformer}. This paper explores the \textit{potential for XAI with natural language inference explanation generation} with a novel deep generative framework. A very recent published paper \cite{cheng2021variational} adopts a similar approach as in this paper, however, the research gap for multiple explanations generation is not explored or discussed. This paper is thus, to the best of our knowledge, the \textit{first work to address the concern on the diversity of human languages} in XAI within the natural language inference task.

\section{Technical Background}

This section provides a brief overview of the Conditional Variational Autoencoder (CVAE), the Transformer architecture, and a description of the data.

\subsection{Conditional Variational Autoencoder}

CVAE \cite{sohn2015learning,larsen2016autoencoding} is an extended version of the deep generative latent variable model (LVM) based on the variational autoencoder (VAE) model \cite{kingma2013auto,rezende2014stochastic}. Both models allow learning rich, nonlinear representations for high-dimensional inputs. When compared with VAE (performing inferences for the latent representation $\boldsymbol{z}$, based on the input $\boldsymbol{x}$, only), CVAE performs inference for the latent representation $\boldsymbol{z}$, based on \textbf{both} the input $\boldsymbol{x}$ and the output $\boldsymbol{y}$, together. CVAE can be considered as a neural network framework based on supervised Neural Variational Inference.


CVAE generally includes two components: an encoder and a decoder. We consider the joint probability distribution and its factorisation, in the form of $\boldsymbol{p_{\theta}(y,z|x) = p_{\theta}(y|z,x) p_{\theta}(z|x)}$ as in \cite{miao2016neural,zhao2017learning,pagnoni2018conditional,gao2019discrete,fang2021transformer}. The encoder $\boldsymbol{p_{\theta}(z|x)}$ takes the observed input $\boldsymbol{x}$ and produces a corresponding latent vector $\boldsymbol{z}$ as the output with parameter $\theta$. The decoder $\boldsymbol{p_{\theta}(y|z,x)}$ takes the observed input $\boldsymbol{x}$ and its corresponding latent vector sample $\boldsymbol{z}$ as the total input and produces an output $\boldsymbol{y}$ with the parameter $\theta$. The latent variable $\boldsymbol{z}$ in the joint probability $\boldsymbol{p_{\theta}(y,z|x)}$ can be marginalised out by taking samples from $\boldsymbol{p(z)}$.

For CVAE, we optimise the following evidence lower bound (ELBO) for the log-likelihood during training:

\begin{equation}
\resizebox{.95\columnwidth}{!}{
$\begin{split}
    \boldsymbol{\log p_{\theta}(y|x) \geq \mathcal{L}(ELBO) = E_{q_{\phi}(z)}[\log p_{\theta}(y|z, x)]  - D_{KL}[q_{\phi}(z|x, y)||p_{\theta}(z|x)]}
\end{split}$}
\label{equ: elbo}
\end{equation}

The first term of ELBO is the reconstruction loss and is measured via cross-entropy matching between predicted versus real targets $\boldsymbol{y}$. The second term is the Kullback–Leibler (KL) divergence between two distributions $\boldsymbol{p_{\theta}(z|x)}$ and $\boldsymbol{q_{\phi}(z|x, y)}$. As the true posterior distribution $\boldsymbol{p_{\theta}(z|x)}$ is intractable to compute, a variational family distribution $\boldsymbol{q_{\phi}(z|x, y)}$ is introduced as its approximation. We consider that both $\boldsymbol{p_{\theta}(z|x)}$ and $\boldsymbol{q_{\phi}(z|x, y)}$ are in the form of isotropic Gaussian distributions, as $\boldsymbol{\mathcal{N}(\mu_{\theta}(x), \textup{diag}(\sigma^{2}_{\theta}(x)))}$ and $\boldsymbol{\mathcal{N}(\mu_{\phi}(x, y), \textup{diag}(\sigma^{2}_{\phi}(x, y)))}$. Our work takes a similar assumption, but the key difference lies in the design of our novel model architectures (section \ref{sec: esnli-main}), together with using the Transformer model \cite{vaswani2017attention} as a building block. 


\subsection{Transformer Architecture}

The Transformer architecture, proposed in \cite{vaswani2017attention}, is the first neural network architecture entirely built upon the self-attention mechanism. It has been used as the main building block for most of the current state-of-the-art models in NLP, such as BERT \cite{devlin2018bert}, GPT3 \cite{brown2020language}, and BART \cite{lewis2019bart}. The Transformer architecture can be divided into three main components: an embedding part, an encoder, and a decoder. 

The embedding part takes the input $x\in R^{s_{1}\times 1}$ in the form of a sequence with length $s_{1}$ and uses an input embedding to create $E(x)\in R^{s_{1}\times E}$, where $E$ is the embedded dimension size. Due to the permutation-invariant self-attention mechanism, \cite{vaswani2017attention} further introduces positional encoding, to encode sequential order information, as $P(x)\in R^{s_{1}\times E}$. The sum of positional encoding and input embedding is used as the final embedding of the input $x$. In \cite{vaswani2017attention}, sine and cosine functions of different frequencies are adopted as positional encoding methods. Further work on large-scale transformers \cite{devlin2018bert,brown2020language,lewis2019bart} use a \textit{learnt positional embedding}, which is what we utilise in this paper. For the encoder and the decoder, we use precisely the same Transformer architecture as in the original paper \cite{vaswani2017attention}. In our experiments, if an encoder and a decoder are used simultaneously, they each have a separate embedding part.


\subsection{Data Description}

Our training data is in the form of $N$ data quadruplets $\{x_{n}^{(p)}, x_{n}^{(h)}, y_{n}^{(l)}, y_{n}^{(e)}\}_{n=1}^{N}$, with each quadruplet consisting of the \textit{premise} (denoted by $x_{n}^{(p)}$), the \textit{hypothesis} (denoted by $x_{n}^{(h)}$) their \textit{associated label} (denoted by $y_{n}^{(l)}$), and \textit{explanation} (denoted by $y_{n}^{(e)}$). For the $n^{th}$ quadruplet, $x_{n}^{(p)} = \{w_{1}^{(p)}, ..., w_{L_{p}}^{(p)} \}$, $x_{n}^{(h)} = \{w_{1}^{(h)}, ..., w_{L_{h}}^{(h)} \}$, $y_{n}^{(l)} = \{w^{(l)} \}$, and $y_{n}^{(e)} = \{w_{1}^{(e)}, ..., w_{L_{e}}^{(e)} \}$ denote the set of $L_{p}$ words from the premise sentence, $L_{h}$ words from the hypothesis sentence, a single word $w^{(l)}$ from the label, and $L_{e}$ words from the explanation sentence, respectively. 

Our validation and testing data are similar to data quadruplets as the training data; however, we have three (${y_{n}^{(e_{1})}}$, ${y_{n}^{(e_{2})}}$ and ${y_{n}^{(e_{3})}}$) instead of one explanation ${y_{n}^{(e)}}$, all created by human experts. During training, we update model parameters based on one explanation ${y_{n}^{(e)}}$ for $n^{th}$ data entry; and during validation and testing, we perform model selection and inference based on the mean average loss of the three explanations (${y_{n}^{(e_{1})}}$, ${y_{n}^{(e_{2})}}$ and ${y_{n}^{(e_{3})}}$). In the following, we omit the data quadruplet index $n$ and use bold characters to represent vector form representations, as $\boldsymbol{x^{(p)}}$, $\boldsymbol{x^{(h)}}$, $\boldsymbol{y^{(l)}}$, and $\boldsymbol{y^{(e)}}$. 

\section{Preliminary Experiments}
\label{sec: preliminary-experiments}


We present two preliminary experiments in this section. We use the architecture setting similar to the \textit{base} version of the Transformer model \cite{vaswani2017attention}, which is a $6$-layer model with $512$ hidden units and $8$ heads for each encoder-decoder network. Based on an inspection of token length statistics (Appendix \ref{sec:appendix-dataset}), we set the maximum length of $25$ for positional encoding. We adopt the pre-processing technique as in \cite{camburu2018snli}. See Appendix \ref{sec:appendix-model-complexity} for a detailed description of all model complexity in this paper.

We generally follow the vocabulary processing steps as in \cite{camburu2018snli} (see detailed pre-processing description in Appendix \ref{sec:appendix-dataset}). We report our quantitative assessment results based on 3 random seeds ($1000$, $2000$, and $3000$), and report the average performance with its standard deviation in parenthesis. Regarding quantitative assessment, we use automatic evaluation metrics (Perplexity and BLEU \cite{papineni2002bleu}) over the entire test data points. Regarding qualitative assessment (Correct$@100$, as in Table \ref{tab: eNLI} and Table \ref{tab: main}), we report results based on the seed $1000$. We adopt the criterion as in \cite{camburu2018snli} and evaluate the Correct$@100$ score based on the first $100$ test examples only\footnote{The score is related to the correctness for generated explanation based on the annotations, details described in Appendix \ref{sec:appendix-correct100-score}.}. For evaluation, the lower the perplexity, the higher the BLEU score and the higher the Correct$@100$ score, the better the model performs.


We use the maximum a posteriori (MAP) estimate decoding for the conditional generation. MAP decoding, whilst not always the optimal choice, has a reasonably good performance and is widely adopted and cheap to compute \cite{eikema2020map}. For the network optimisation, we use Adam \cite{kingma2014adam} as our optimiser with default hyperparameters ($\beta_{1}=0.9$, $\beta_{2}=0.999$, $\epsilon=1e-8$). We conduct all the experiments with a batch size of $16$ and a learning rate of $1e-5$ for a total of $10$ epochs on a machine with Ubuntu $20.04$ operating system and a GTX $2080$Ti GPU.

\subsection{Architecture Selection and Spurious Correlation}
\label{sec: snli-experiments}

In the first experiment, we answer two questions: \textbf{Q(i)} \textit{What is a good Transformer model architecture choice for the e-SNLI text classification task?} \textbf{Q(ii)} \textit{How easily can a Transformer model pick up the spurious correlation, when only a hypothesis sentence is observed?}

\begin{figure}[h]
\centering
\includegraphics[width=0.50\columnwidth]{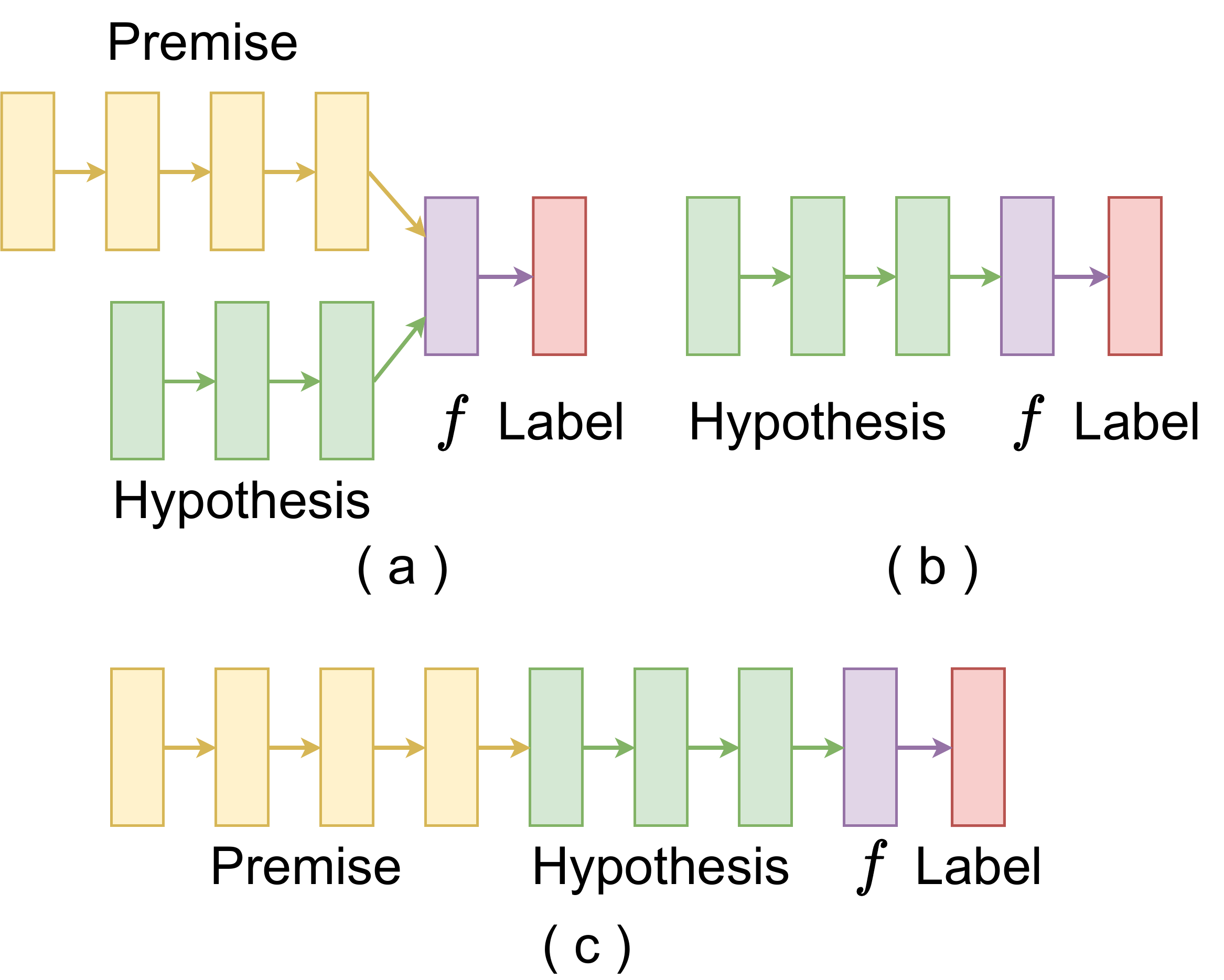}
\caption{Graphical overview of architectures used in section \ref{sec: snli-experiments}. (a) Separate Transformer Encoder; (b) Premise Agnostic Encoder; and (c) Mixture Transformer Encoder.}
\label{fig: snli-architecture}
\end{figure}

\begin{table}[h]
\smallskip
\centering
\caption{Architecture Selection and Spurious Correlation Experiments.}
\resizebox{.60\columnwidth}{!}{
\smallskip\begin{tabular}{ll}
\hline
Model & Accuracy (\%) \\
\hline
Separate Transformer Encoder & $73.97$ ($0.34$)\\
Mixture Transformer Encoder & $78.98$ ($1.44$)\\
\hline
Premise Agnostic Encoder & $65.43$ ($0.72$)\\
\hline
\end{tabular}}
\label{tab: SNLI}
\end{table}

To answer \textbf{Q(i)}, we experiment on two candidate model architectures: \textit{(1)} \textit{Separate Transformer Encoder}: an architecture with two separate encoders, one each for the premise and hypothesis sentences, respectively (Fig. \ref{fig: snli-architecture}a). \textit{(2)} \textit{Mixture Transformer Encoder}: an architecture with a mixture encoder for both premise and hypothesis sentence together (Fig \ref{fig: snli-architecture}c). We choose these two candidates for the following reasons: the first candidate architecture is widely adopted in early NLI literature \cite{parikh2016decomposable,chen2017neural,gong2017natural},  where $f$ refers to algorithmic operations (identity, subtraction, multiplication) as in \cite{conneau2017supervised}. The latter candidate architecture is adopted by the BERT model \cite{devlin2018bert}, where $f$ refers to an affine transformation operation and has achieved state-of-the-art performance for NLI tasks. To answer \textbf{Q(ii)}, we perform the premise-agnostic prediction experiment on the \textit{Premise Agnostic Encoder} model (Fig \ref{fig: snli-architecture}b), where $f$ refers to an affine transformation operation.

For the above two experiments, results are presented in Table \ref{tab: SNLI}. For the \textit{Separate Transformer Encoder}, we use the encoder outputs at two separate '$<bos>$' positions for algorithmic operations (identity, subtraction, and multiplication). For \textit{Mixture Transformer Encoder} and \textit{Premise Agnostic Encoder}, we use the output at the first '$<bos>$' position. We apply an affine transformation operation for predicting the label. The results suggest the \textit{Mixture Transformer Encoder} outperforms the \textit{Separate Transformer Encoder}, in a statistically significant way ($p <.05$; Wilcoxon test). The \textit{Premise Agnostic Encoder} achieves $82.84\%$ (based on $65.43/78.98$) of the \textit{Mixture Transformer Encoder} performance, suggesting that Transformer models tend to capture spurious correlations very easily for the NLI label prediction task.

\subsection{Premise-Agnostic and Full Generation}
\label{sec: esnli-agnositc}

In the second experiment, we address two further questions: \textbf{Q(iii)} \textit{Is providing explanations as output reducing the impact of spurious correlation in a Transformer model, compared to predicting the label only?} \textbf{Q(iv)} \textit{How much better are explanations based on premise and hypothesis together, instead of hypothesis-only?}

\begin{figure}[h]
\centering
\includegraphics[width=0.45\columnwidth]{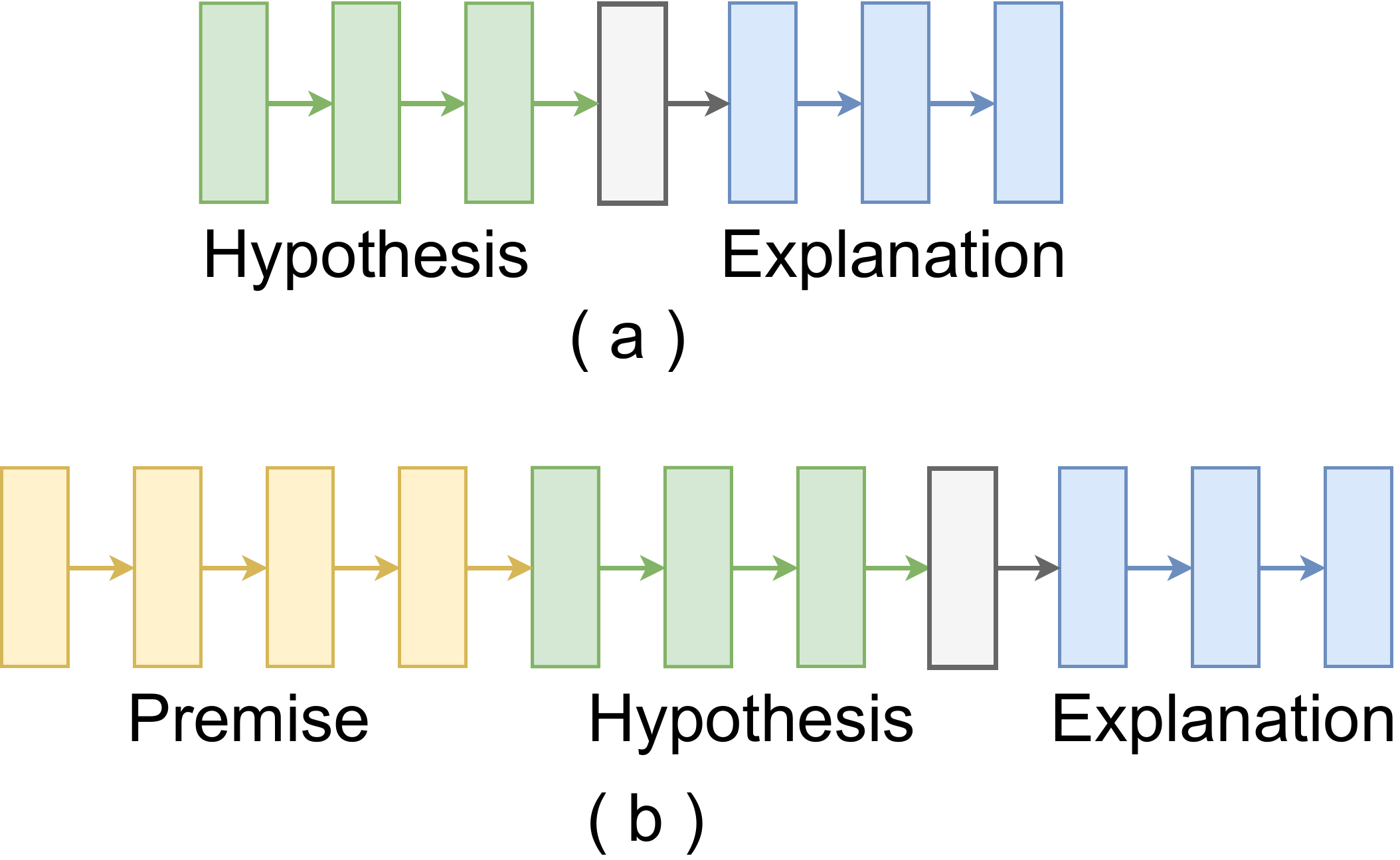}
\caption{Graphical overview of architectures used in section \ref{sec: esnli-agnositc}. (a) Agnostic Generation; (b) Full Generation.}
\label{fig: esnli-seq2seq}
\end{figure}

\begin{table}[h]
\smallskip
\centering
\caption{Premise Agnostic Generation Experiments.}
\resizebox{.85\columnwidth}{!}{
\smallskip\begin{tabular}{lccc}
\hline
Model & Perplexity & BLEU & Correct$@100$\\
\hline
Agnostic Generation & $7.66$ ($0.03$) & $25.74$ ($0.8$) & $42.87$\\
Full Generation & $5.53$ ($0.05$) & $33.14$ ($0.5$) & $57.45$\\
\hline
\end{tabular}}
\label{tab: eNLI}
\end{table}

To answer \textbf{Q(iii)}, we follow and extend the \textit{'PremiseAgnostic'} experiment \cite{camburu2018snli}. We use the model architecture shown in Fig. \ref{fig: esnli-seq2seq}a, and we are interested in evaluating how well the model can generate an explanation from the premise-agnostic scenario (only premise observed). To answer \textbf{Q(iv)}, we implement the seq2seq framework \cite{sutskever2014sequence} with the Transformer architecture. We compare the agnostic generation scenario with the full generation scenario (both premise and hypothesis observed, as shown in Fig. \ref{fig: esnli-seq2seq}b). 


Our results, presented in Table \ref{tab: eNLI}, suggest that the agnostic generation significantly reduces ($p <.05$; Wilcoxon test) the ability to generate correct explanations, with only $72.19\%$ (based on $5.53/7.66$) for perplexity, $77.67\%$ (based on $25.74/33.14$) for the BLEU score, and $74.62\%$ (based on $42.87/57.45$) for the Correct$@100$ score (compared to $82.84\%$ in section \ref{sec: snli-experiments}). 





\section{Proposed Deep Generative XAI Framework}
\label{sec: esnli-main}

In this section, we explain in detail our novel framework, \textbf{INTERACTION} - (expla\underline{\textbf{I}}n a\underline{\textbf{N}}d predic\underline{\textbf{T}} th\underline{\textbf{E}}n que\underline{\textbf{R}}y with contextu\underline{\textbf{A}}l \underline{\textbf{C}}ondi\underline{\textbf{T}}ional var\underline{\textbf{I}}ational aut\underline{\textbf{O}}-e\underline{\textbf{N}}coder). Our framework presents explanation in two steps: (step one) \textit{Explanation and Label Prediction}; and (step two) \textit{Diverse Evidence Generation}. We present a workflow diagram for our framework in Fig. \ref{fig: framework}, which consists of four components as follows.


\subsection{Neural Encoder}
Given a pair of premise $\boldsymbol{x^{(p)}}$ and hypothesis $\boldsymbol{x^{(h)}}$, with their associated explanation $\boldsymbol{y^{(e)}}$, the encoder network outputs two sequences of representations:

\begin{equation}
\resizebox{.40\columnwidth}{!}{
$\begin{split}
    \boldsymbol{x_{h} = Encoder([x^{(p)};x^{(h)}])} \\ \boldsymbol{y_{h} = Encoder([y^{(e)}])}
\end{split}$}
\label{equ: encoder}
\end{equation}

Here $\boldsymbol{Encoder}$ refers to the \textit{Transformer Mixture Encoder}, which is selected based on experiments in section \ref{sec: snli-experiments}. $\boldsymbol{x_{h}}$ is the contextual representations for the premise $\boldsymbol{x^{(p)}}$ and hypothesis $\boldsymbol{x^{(h)}}$ pair. $\boldsymbol{y_{h}}$ is the contextual representation for explanation $\boldsymbol{y^{(e)}}$. We share the same encoder network parameters for producing $\boldsymbol{x_{h}}$ and $\boldsymbol{y_{h}}$. $\boldsymbol{x_{h}}$ has the same sequence length as the sum of premise and hypothesis length. $\boldsymbol{y_{h}}$ has the same sequence length as the explanation length. $\boldsymbol{[a;b]}$ refers to the concatenation operation of vectors $\boldsymbol{a}$ and $\boldsymbol{b}$.

\subsection{Neural Inferer}

The neural inferer can be divided into two separate components: the prior and the posterior networks. As determined by the ELBO equation \ref{equ: elbo}, the parameters of the prior are computed by the prior network, which only takes the inputs: $\boldsymbol{x^{(p)}}$ and $\boldsymbol{x^{(h)}}$. The posterior parameters are determined from both inputs and outputs: $\boldsymbol{x^{(p)}}$, $\boldsymbol{x^{(h)}}$ and $\boldsymbol{y^{(e)}}$.


\subsubsection{Contextual Convolutional Neural Encoder}
\label{sec: contextual-cnn}

Before introducing the neural prior and posterior, we first present our novel approach of dealing with various lengths of output from the Transformer encoder. We first adopt the 2d-convolution operations (over the sequence length and hidden dimension) as in \cite{kim-2014-convolutional} and apply it directly to the encoded outputs $\boldsymbol{x_{h}}$ and $\boldsymbol{y_{h}}$. For the convolution operations, we use learnable filters with size of $1$, $2$, and $3$ to represent '\textit{unigram}', '\textit{bigram}', and '\textit{trigram}' contextual information from the sequences. Then, we use a max-pooling operation over each filter output, to alleviate various sequence-length issues and concatenate them as one single output vector. Finally, we apply an affine transformation on the output vector and return the original vector dimension, but with a sequence length of $1$. We name the whole set of operations here \textbf{c}ontextual c\textbf{o}nvolutional \textbf{n}eural en\textbf{coder} (denoted in short as $\boldsymbol{Concoder}$).

In contrast, a standard CVAE model uses a fixed position from the sequence instead, to handle various sequence-length issues. We implement a standard CVAE with the $<bos>$ position output as the final output, denoted as \textit{CVAE Generation}. We use this as a comparison with our novel solution ($\boldsymbol{Concoder}$), denoted as \textit{ConCVAE Generation} (with results shown in Table \ref{tab: main}).

\subsubsection{Neural Prior}

The prior distribution, denoted as:

\begin{equation}
\resizebox{.50\columnwidth}{!}{$
    \boldsymbol{p_{\theta}(z|x) = \mathcal{N}(z|\mu_{\theta}(x), \textup{diag}(\sigma^{2}_{\theta}(x)))}$}
\end{equation}

$\boldsymbol{p_{\theta}(z|x)}$ is an isotropic multivariate Gaussian with mean and variance matrices parameterised by neural networks. With variable-length sentence as input, we first use a contextual convolutional neural network, introduced in section \ref{sec: contextual-cnn}, to retrieve a fixed output $\boldsymbol{x_{c}}$. Then, we apply two additional affine transformations, $\boldsymbol{f_{1}}$ and $\boldsymbol{f_{2}}$, to parameterise the mean and variance matrices for the neural prior. The $\boldsymbol{tanh()}$ function here introduces additional non-linearity and also contributes to numerical stability during parameter optimisation. Thus:

\begin{equation}
\resizebox{.40\columnwidth}{!}{
$
\begin{split}
    \boldsymbol{x_{c} = Concoder([x_{h}])} \\  \boldsymbol{\mu_{\theta}=f_{1}([x_{c}])} \\ \boldsymbol{\log\sigma_{\theta}=tanh(f_{2}([x_{c}]))}
\end{split}
$}
\label{equ: prior}
\end{equation}

\subsubsection{Neural Posterior}

During training, the latent variable will be sampled from the posterior distribution:

\begin{equation}
\resizebox{.60\columnwidth}{!}{$
    \boldsymbol{q_{\phi}(z|x, y) = \mathcal{N}(z|\mu_{\phi}(x, y), \textup{diag}(\sigma^{2}_{\phi}(x, y)))}$}
\end{equation}

$\boldsymbol{q_{\phi}(z|x, y)}$ is also an isotropic multivariate Gaussian with mean and variance matrices parameterised by neural networks. However, the parameters are inferred based on both inputs and outputs. We use the same $\boldsymbol{Concoder}$ network to handle the various lengths of inputs and outputs ($\boldsymbol{x^{(p)}}$, $\boldsymbol{x^{(h)}}$, and $\boldsymbol{y^{(e)}}$).  As for the neural prior, we apply two additional affine transformations, $\boldsymbol{f_{3}}$ and $\boldsymbol{f_{4}}$, to parameterise the mean and variance matrices. Thus:

\begin{equation}
\resizebox{.40\columnwidth}{!}{
$\begin{split}
    \boldsymbol{y_{c} = Concoder([y_{h}])} \\  \boldsymbol{\mu_{\phi}=f_{3}([x_{c}; y_{c}])} \\ \boldsymbol{\log\sigma_{\phi}=tanh(f_{4}([x_{c}; y_{c}]))}
\end{split}
$
}
\label{equ: posterior}
\end{equation}

\subsection{Neural Decoder}

The decoder models the probability of the explanation $\boldsymbol{y^{(e)}}$ in an auto-regressive manner, given the predicted label $\boldsymbol{y_{p}}$, the encoded premise and hypothesis pair $\boldsymbol{x_{h}}$, and the latent vector $\boldsymbol{z}$. We obtain the explanation sequence via:

\begin{equation}
\resizebox{.40\columnwidth}{!}{
$\begin{split}
    \boldsymbol{y^{(e)} = Decoder([z; x_{(h)}])}
\end{split}$}
\label{equ: decoder}
\end{equation}

Here, $\boldsymbol{Decoder}$ refers to the Transformer decoder. Given an explanation with a total sequence length of $T$, at time step $j$ ($j<T$), it produces the $j^{th}$ word with a softmax selection from the vocabulary based on all the past $j-1$ words.

\begin{figure}[t]
\centering
\includegraphics[width=0.50\columnwidth]{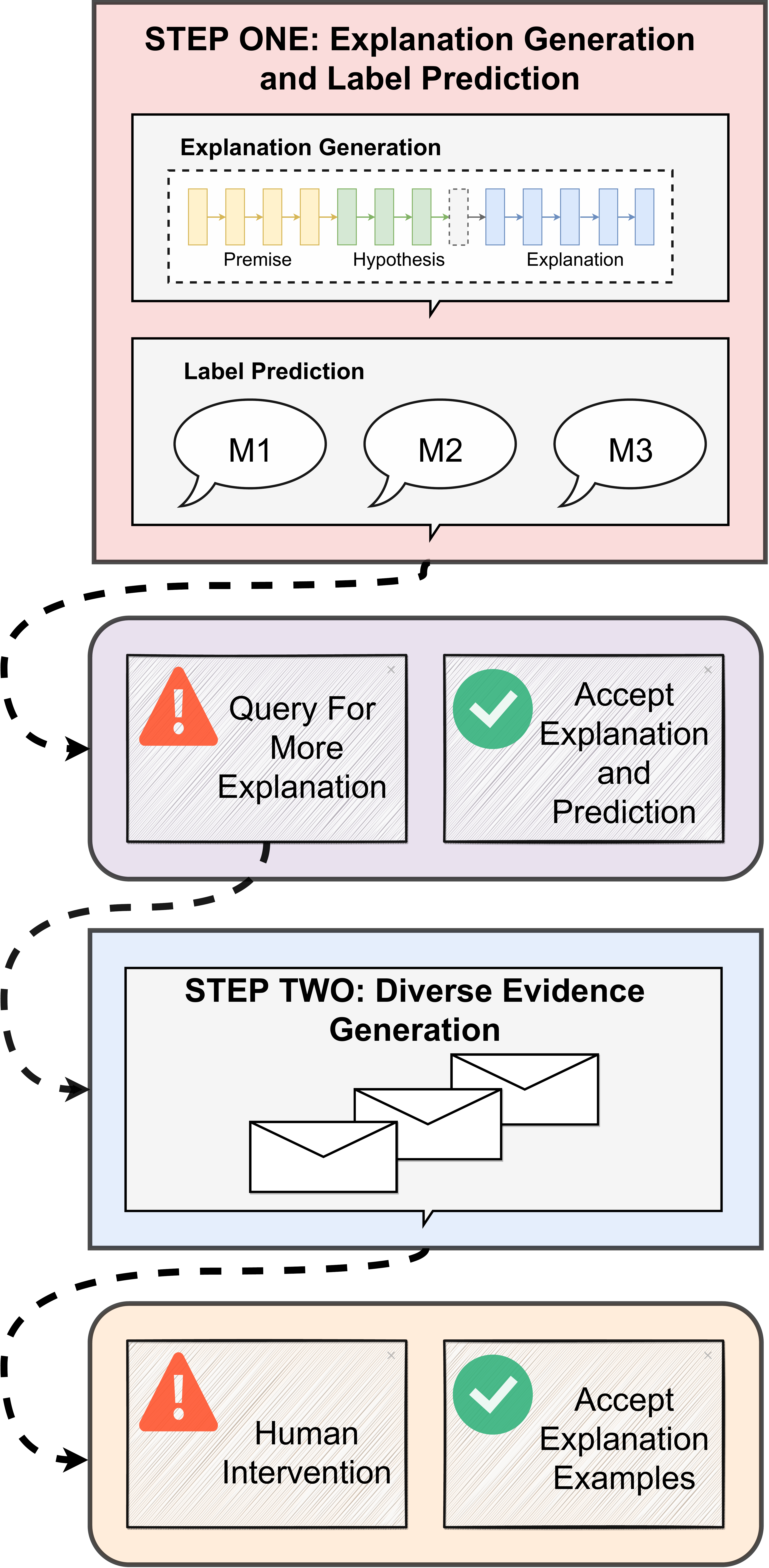}
\caption{Graphical overview of our framework, \textbf{INTERACTION}, introduced in section \ref{sec: esnli-main}.}
\label{fig: framework}
\end{figure}

\subsection{Neural Predictor}
\label{sec: neural-predictor}

In our novel \textbf{INTERACTION} framework, the label can be predicted based on one of the three options: \textit{(i)} \textbf{M1 Model}: predicted based on the premise and hypothesis only, \textit{(ii)} \textbf{M2 Model}: predicted based on the explanation only, and \textit{(iii)} \textbf{M3 Model}: predicted based on the premise, hypothesis, and explanation all together. With the Transformer architecture, we first concatenate the vector outputs of the information at each first '$<bos>$' position into a single vector for each model. Then we apply an affine transformation operation $\boldsymbol{f}$ to the concatenated vector. We jointly train the neural predictor together with the generative model \textit{ConCVAE}. We compare the performance of these three models in our experiments (Table \ref{tab: main}).

\section{Experiments}
\label{sec: main-experiment}

In this section, to evaluate our proposed framework \textbf{INTERACTION}, we conduct experiments comparing with our baseline models.

\begin{table*}[t]
\smallskip
\small
\centering
\caption{XAI with natural language processing Results ( `$--$' refers to results not applicable).}
\resizebox{.75\textwidth}{!}{
\smallskip\begin{tabular}{lcccc}
\hline
Model & Label Accuracy & Perplexity & BLEU & Correct$@100$\\
\hline
Premise Agnostic Encoder (lower bound) & $65.43$ ($0.72$) & $--$ & $--$ & $--$ \\
Mixture Transformer Encoder (predictive model baseline) & $\textbf{78.98}$ ($1.44$) & $--$ & $--$ & $--$ \\
\hline
Full Generation (generative model baseline, non-probabilistic) & $--$ & $\textbf{5.53}$ $(0.05)$ & $\textbf{33.14}$ $(0.50)$ & $\textbf{57.45}$\\
CVAE Generation (generative model baseline, probabilistic) & $--$ & $7.58$ $(0.27)$ & $25.70$ ($1.04$) & $43.04$\\
ConCVAE Generation (our model, probabilistic) & $--$ & $5.69$ $(\textbf{0.03})$ & $32.74$ $(\textbf{0.09})$ & $55.27$\\
\hline
INTERACTION M1 (our model) & $\textbf{83.42 (0.31)}$ & $6.73 (0.16)$ & $30.46 (0.33)$ & $47.04$\\
INTERACTION M2 (our model) & $73.73 (1.54)$ & $\textbf{5.75 (0.01)}$ & $32.68 (0.64)$ & $52.29$\\
INTERACTION M3 (our model) & $79.85 (0.35)$ & $5.93 (0.02)$ & $\textbf{32.70 (0.28)}$ & $\textbf{58.06}$\\
\hline
\end{tabular}}
\label{tab: main}
\end{table*}

\begin{table*}[h]
\smallskip
\small
\centering
\caption{Selected diverse evidence generation examples.}
\resizebox{.70\linewidth}{!}{
\smallskip\begin{tabular}{lp{0.99\linewidth}}
\hline
\hline
Test Data Number & 29\\
\hline
Premise & a \textbf{couple} walk hand in hand down a street . \\
\hline
Hypothesis & the \textbf{couple} is \textbf{married} .\\
\hline
Explanation & just because the couple is hand in hand does n't mean they are married .\\
\hline
Generated Explanation 1 & not all couple walking down street are married .\\
\hline
Generated Explanation 2 & not all couple in hand is married .\\
\hline
Generated Explanation 3 & not all couples are married .\\
\hline
\hline
Test Data Number & 50\\
\hline
Premise & a little boy in a gray and white striped sweater and tan pants is \textbf{playing} on a piece of \textbf{playground equipment} .\\
\hline
Hypothesis & the boy is \textbf{sitting on} the school bus on his \textbf{way home} .\\
\hline
Explanation & the boy is either playing on a piece of playground equipment or sitting on the school bus on his way home .\\
\hline
Generated Explanation 1 & the boy can not be playing on a playground and sitting on his way home at the same time .\\
\hline
Generated Explanation 2 & the boy can not be playing on a playground and sitting on his way home simultaneously .\\
\hline
Generated Explanation 3 & the boy can not be playing on a playground and sitting on the bus at the same time .\\
\hline
\hline
Test Data Number & 64\\
\hline
Premise & people jump over \textbf{a mountain} crevasse on a rope .\\
\hline
Hypothesis & people are jumping \textbf{outside} .\\
\hline
Explanation & the jumping over the mountain crevasse must be outside .\\
\hline
Generated Explanation 1 & people jump over a mountain so they must be outside .\\
\hline
Generated Explanation 2 & a mountain is outside .\\
\hline
\hline
Test Data Number & 77\\
\hline
Premise & a man in a black shirt is \textbf{looking at} a bike in a workshop .\\
\hline
Hypothesis & a man is deciding which bike \textbf{to buy}\\
\hline
Explanation & just because the man is looking at a bike does n't mean he is deciding which bike to buy .\\
\hline
Generated Explanation 1 & just because a man is looking at a bike in a workshop does n't mean he is deciding to buy .\\
\hline
Generated Explanation 2 & just because a man is looking at a bike in a workshop does n't mean he is deciding what to buy .\\
\hline
\hline
\end{tabular}}
\label{tab: interpolation}
\end{table*}

\subsection{Baseline Models}
We define two types of baseline models: \textit{generative model} and \textit{predictive model}. We consider the following works as baseline models:
\begin{itemize}
  \item seq2seq (\textit{generative model}, our implementation): a sequence-to-sequence learning framework developed by \cite{sutskever2014sequence}. We implement it with the Transformer architecture and present the results as \textit{Full Generation} in Table \ref{tab: main}.
  \item CVAE (\textit{generative model}, our implementation): a strong probabilistic conditional generation framework introduced by \cite{sohn2015learning, larsen2016autoencoding}. We implement it with the Transformer architecture and present the results as \textit{CVAE Generation} in Table \ref{tab: main}.
  \item Transformer (\textit{predictive model}, our implementation): a very strong baseline model for NLI task developed by \cite{vaswani2017attention}. We present the results as \textit{Mixture Transformer Encoder} in Table \ref{tab: main}.
\end{itemize}

\subsection{Experiment Setup}

To evaluate the explanation generative model of our \textbf{INTERACTION} framework, we implement our novel \textit{ConCVAE} model and use the MAP decoding over the latent variable during both training and testing to generate a single explanation. For label prediction task, we implement the \textbf{INTERACTION} \textbf{M1}, \textbf{M2}, and \textbf{M3} models (as in section \ref{sec: neural-predictor}), and compare their performance with our predictive and generative baseline models. Regarding network architectures, vocabulary, and training, we use the same experimental setting as in section \ref{sec: preliminary-experiments}.

\subsection{Diverse Evidence Generation via Interpolation}
\label{sec: latent-space-interpolation}

We present a study on the generation of diverse evidence to support explanation, as in \textit{step two} from Figure \ref{fig: framework}. To generate multiple explanations, we perform posterior analysis over the latent space. We choose to linearly interpolate the isotropic multivariate Gaussians over its $95.44\%$ region (left and right of 2$\sigma$ from $\mu$). This interpolation produces $5$ samples calculated based on the $\mu-2\sigma$, $\mu-\sigma$, $\mu$, $\mu+\sigma$, and $\mu+2\sigma$ coordinates. Examples of interpolation results from the \textit{ConCVAE Generation} experiment are presented in Table \ref{tab: interpolation} and we only show the examples which are different. 



\section{Results and Discussions}

\subsection{Explanation Generation Only}

The main results are presented in Table \ref{tab: main}. For the explanation generation evaluation, we first compare a deep generative model (\textit{CVAE Generation}) with a standard neural network model (\textit{Full Generation}). The results suggest that the \textit{Full Generation} model performs better, as the perplexity is reduced by ($7.58 - 5.53 = 2.05$), the BLEU score increases by ($33.14 - 25.70 = 7.4\%$), and the Correct$@100$ score increases ($57.45 - 43.04 = 14.4$). All the results here are statistically significant ($p < .05$) based on the Wilcoxon signed-rank test. However, deep generative models, such as \textit{CVAE Generation}, allow generating multiple explanations via a posterior analysis over the latent space, as shown in section \ref{sec: latent-space-interpolation}. With our novel contextual deep generative model \textit{ConCVAE}, we achieve competitive performance with the \textit{Full Generation} model, evidenced in both quantitative (perplexity, BLEU score) and qualitative (Correct $@100$) results.

\subsection{Explanation Generation and Label Prediction}

We implement three variants of our \textbf{INTERACTION} framework (\textbf{M1}, \textbf{M2} and \textbf{M3}) to perform generation and prediction simultaneously. Regarding label prediction, results suggest that generating a valid explanation from the premise and hypothesis sentence-pair allows the encoder to better understand the semantics of the words and hence further enhances the accuracy of prediction. This leads to a boost in prediction performance ($83.42\%$ for \textbf{M1} and $79.85\%$ for \textbf{M3}), compared to the \textit{Mixture Transformer Encoder} ($78.98\%$), with the same number of parameters. However, with \textbf{M1}, a significant improvement in classification accuracy results in the worst generation quality (based on Correct$@100$) among all three models. Additionally, as shown in \textbf{M2} model, the label prediction accuracy is the worst when using explanation only. This could potentially be explained since only $52.29\%$ of the explanations are considered as correct (based on Correct$@100$).

Regarding explanation generation, we observe that the \textbf{M3} model achieves competitive results for the quantitative assessment (perplexity and BLEU) as the \textit{Full Generation} model. Additionally, it achieves the best performance in qualitative assessment (Correct$@100$) amongst all models. The results from Table \ref{tab: main} suggest that label prediction and explanation generation can complement each other and hence enhance the importance of \textit{XAI with natural languages} in practice. When choosing amongst these three models: for the prediction performance,the \textbf{M1} model fits the best; however, for the generation performance, the \textbf{M3} model is preferable.

\subsection{Diversity of Explanation}

The main contribution in this paper is to build a model (\textbf{INTERACTION}) capable of providing multiple explanations, reflecting the diversity in natural languages. The motivation is that a natural language usually works in a way such that humans often provide more than one explanation for their actions, and hence may find systems that reply 'monosyllabically', or too briefly, potentially frustrating, or even non-informative \cite{miller2019explanation,zhou2021paraphrase}. Still, our approach raises other questions, e.g., do humans have enough time to read multiple explanations? How do they pick the best or most faithful one? In a recent paper \cite{cho2019paraphrase}, the authors propose first to generate multiple paraphrases and then select the most faithful one. In our paper (Fig. \ref{fig: framework}), we alternatively select the most faithful one based on MAP decoding in \textit{step one} (the maximum likelihood for data), then provide multiple explanations in \textit{step two}. The richness and diversity of the generation of multiple explanations can be observed in Table \ref{tab: interpolation} (e.g., for test data number $29$ and $64$). In practice, the MAP decoding might not offer the best results; however, it is a faithful response from the model, given the context of using a `data-driven' approach with deep learning.


\section{Conclusion}

Here, we have presented \textbf{INTERACTION}, a novel deep generative XAI framework, with explanations in two steps: (1) Explanation and Label Prediction; and (2) Diverse Evidence Generation. INTERACTION is the first study which, to the best of our knowledge, \textit{addresses the concern on the diversity of human languages} in XAI, within the natural language processing task. INTERACTION achieves competitive or better performance against state-of-the-art baseline models on both generation (4.7\% improvement in BLEU) and prediction (4.4\% improvement in accuracy) tasks. We observe that label prediction and explanation generation can complement each other, which further confirms the benefits of \textit{XAI with natural languages} research in practice.

\bibliographystyle{IEEEtran}
\bibliography{IEEEabrv,ieee}

\appendices

\section{Dataset Statistics}
\label{sec:appendix-dataset}

\begin{table}[h]
\smallskip
\centering
\caption{Token length statistics for the e-SNLI dataset, all numbers round to integer.}
\resizebox{.60\columnwidth}{!}{
\smallskip\begin{tabular}{lccccc}
\hline
Model & Mean & Median & Standard Deviation & Min & Max \\
\hline
Premise & 17 & 15 & 7 & 4 & 84 \\
Hypothesis & 11 & 10 & 4 & 3 & 64 \\
Explanation & 16 & 15 & 7 & 2 & 189 \\
\hline
\end{tabular}}
\label{appendix: dataset}
\end{table}


\section{Qualitative Evaluation}
\label{sec:appendix-correct100-score}

We calculate the qualitative assessment score, Correct$@100$, as suggested in \cite{camburu2018snli}: we manually grade the correctness of first $100$ test examples, each with a score between $0$ (incorrect) and $1$ (correct) and give partial scores of $k/n$ if only $k$ out of $n$ required arguments were mentioned. The require arguments are publicly available on GitHub\footnote{https://github.com/OanaMariaCamburu/e-SNLI/tree/master/dataset} and we take the mean average of three annotations as the final score.

\section{Model Complexity}
\label{sec:appendix-model-complexity}

We present the model complexity in Table \ref{tab: model-complexity}, with separate counts for prediction, generation and total network components, the one with the `$--$' mark is denoted as not applicable.

\begin{table}[h]
\smallskip
\small
\centering
\caption{Number of parameters for each model, with separate counts for prediction and generation component.}
\resizebox{.70\columnwidth}{!}{
\smallskip\begin{tabular}{lccccc}
\hline
Model & Prediction & Generation & Total\\
\hline
Separate Transformer Encoder & $48.6$M & -- & $48.6$M \\
Mixture Transformer Encoder & $24.3$M & -- & $24.3$M \\
Premise Agnostic Encoder & $24.3$M & -- & $24.3$M \\
\hline
Agnostic Generation & -- & $63.6$M & $63.6$M \\
Full Generation & -- & $63.6$M & $63.6$M \\
CVAE Generation & -- & $65.9$M & $65.9$M \\
ConTrCVAE Generation & -- & $68.3$M & $68.3$M \\
\hline
INTERACTION M1 & $24.3$M & $68.3$M & $68.3$M \\
INTERACTION M2 & $24.3$M & $68.3$M & $68.3$M \\
INTERACTION M3 & $24.3$M & $68.3$M & $68.3$M \\
\hline
\end{tabular}}
\label{tab: model-complexity}
\end{table}


\end{document}